\documentclass{article}

\usepackage{PRIMEarxiv}

\usepackage[utf8]{inputenc} 
\usepackage[T1]{fontenc}    
\usepackage{hyperref} 
\usepackage{url}            
\usepackage{booktabs}       
\usepackage{amsfonts}       
\usepackage{nicefrac}       
\usepackage{microtype}      
\usepackage{lipsum}
\usepackage{fancyhdr}       
\usepackage{graphicx}       
\graphicspath{{media/}}     
\usepackage{amsmath}
\usepackage{amssymb}
\usepackage{subfigure}

\graphicspath{{figures/}}

\title{Training Invertible Neural Networks as Autoencoders
\thanks{\textit{\underline{Citation}}: 
\textbf{Nguyen, TG.L., Ardizzone, L., Köthe, U. (2019). Training Invertible Neural Networks as Autoencoders. In: Fink, G., Frintrop, S., Jiang, X. (eds) Pattern Recognition. DAGM GCPR 2019. Lecture Notes in Computer Science, vol 11824. Springer, Cham. https://doi.org/10.1007/978-3-030-33676-9\_31}} 
}

\author{
  The-Gia Leo Nguyen  \\
  Computer Vision and Learning Lab, IWR \\
  Heidelberg University \\
  \texttt{leo.nguyen@iwr.uni-heidelberg.de} \\
  \And
  Lynton Ardizzone \\
  Computer Vision and Learning Lab, IWR \\
  Heidelberg University \\
  \texttt{lynton.ardizzone@iwr.uni-heidelberg.de} \\
  \And
  Ullrich Köthe \\
  Computer Vision and Learning Lab, IWR \\
  Heidelberg University \\
  \texttt{ullrich.koethe@iwr.uni-heidelberg.de}
}

\begin{document}
\maketitle

\begin{abstract}
	Autoencoders are able to learn useful data representations in an unsupervised matter and have been widely used in various machine learning and computer vision tasks. In this work, we present methods to train Invertible Neural Networks (INNs) as (variational) autoencoders which we call \textit{INN (variational) autoencoders}. Our experiments on MNIST, CIFAR and CelebA show that for low bottleneck sizes our INN autoencoder achieves results similar to the classical autoencoder. However, for large bottleneck sizes our INN autoencoder outperforms its classical counterpart. Based on the empirical results, we hypothesize that INN autoencoders might not have any intrinsic information loss and thereby are not bounded to a maximal number of layers (depth) after which only suboptimal results can be achieved. \footnote{Code available at https://github.com/Xenovortex/Training-Invertible-Neural-Networks-as-Autoencoders.git} 
\end{abstract}

\keywords{Machine Learning \and Generative Models \and INN \and Normalizing Flows \and Autoencoder}

\section{Introduction}
\label{sec:introduction}
	
	In machine learning and computer vision, CNNs have been proven to be effective for various tasks, such as object detection \cite{object_detection}, image captioning \cite{image_captioning},  semantic segmentation \cite{semantic_segmentation}, object recognition \cite{object_recognition} or scene classification \cite{scene_classification}. However, all these approaches are based on supervised learning methods and require tremendous amounts of manually labeled data. This can be a limitation, since labeling images commonly involves human effort, which is impractical, expensive and not realizable on a large scale. 
	
	As a result, current research has moved more towards unsupervised learning methods. In particular, autoencoders and VAEs (see \cite{variational_inference,tutorial_vae,vae}) play a fundamental role in learning encoded representations of the data in an unsupervised manner. 
	
	The idea of Invertible Neural Networks (INNs) goes back to the works of Dinh et al. \cite{nice,real_nvp}, which introduced tractable invertible coupling layers. Since then, the research of INNs has seen some relevant advances in further understanding the characteristics of INNs, their relation to classical models and applying INNs to common deep learning tasks. Recent works on INNs include the RevNet from Gomez et al. \cite{rev_residual_net}. They show that RevNets can achieve the same performance as traditional ResNets \cite{resnet} of equal size on classification tasks. Jacobsen et al. \cite{i_revnet} introduced the iRevNet, a fully invertible network that can reproduce the input based on the output of the last layer. They additionally show that the lack of information reduction does not affect the performance negatively. Impressive results have been achieve with Glow-type networks as proposed by Kingma et al. \cite{glow}. The application of INNs on generative tasks has been done by Danihelka et al. \cite{gan_real_nvp}, Schirrmeister et al. \cite{generative_reversible_net} and Grover et al. \cite{flow_gan}. Ardizzone et al. \cite{lynton} have successfully applied INNs on real world problems while proposing their own version of INNs, which allow for bi-directional training. Jacobson et al. \cite{i_revnet} and Grathwohl et al. \cite{scale_revnet} have observed similar behaviors between ResNets and INNs such as iRevNets and Glow-type networks. Leveraging the similarities between ResNets and INNs, Behrmann et al. \cite{invertible_ResNet} were able to train the standard ResNet architecture to learn an invertible bijective mapping by adding a normalization step. A ResNet trained this way can be used for classification, generation and density estimation. Much of how INNs learn and their relation to traditional neural networks are still unknown and subject to current research endeavors. However, recent works on excessive invariance have given some insights in further understanding INNs such as \cite{behrmann_excess,gilmer_excess,jacobsen_excess}.  	
	
	In this work, we propose methods to train INNs as (variational) autoencoders which we call \textit{INN (variational) autoencoder}. We compared their performance to conventional autoencoders for different bottleneck sizes on MNIST, CIFAR and CelebA. For all experiments, we made sure that the INN autoencoders and their classical counterparts have similar number of trainable parameters, where all classical models were given an advantage in number of trainable parameters.  
	
	Our main contributions are:
	\begin{itemize}
		\item
		We propose a method to train INNs as (variational) autoencoders.
		\item
		We compare the performance of INN autoencoders and classical autoencoders for different bottleneck sizes on MNIST, CIFAR and CelebA.
		\item
		We demonstrate through experiments that INN autoencoders can achieve similar or better reconstruction results than classical autoencoders with comparable number of trainable parameters.
		\item 
		We show that the architecture restrictions on INN autoencoders to ensure invertibility does not negatively affect the performance, while the advantages of INNs are still preserve (such as tractable Jacobian for both forward and inverse mapping as well as explicit computation of posterior probabilities).
		\item
		We provide an explanation for the saturation in reconstruction loss for large bottleneck sizes in classical autoencoders.
		\item
		Based on our experimental results, we propose the hypothesis that INNs might not have any intrinsic information loss and thereby are not bounded to a maximal number of layers (depth) after which only suboptimal results can be achieved.
	\end{itemize}

\section{Training INNs as Autoencoders}

\subsection{Invertible Neural Network (INN)}
	\begin{figure}
		\centering
		\includegraphics[width=0.9\linewidth]{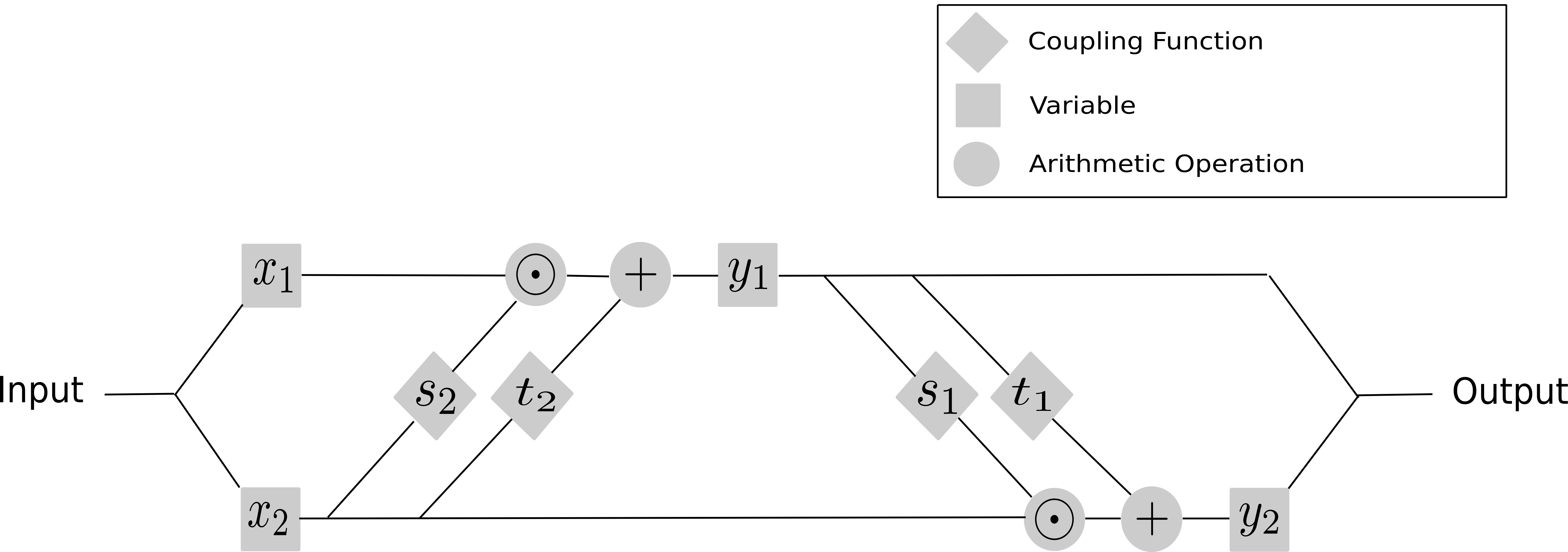}
		\caption{Visualization of the Invertible Coupling Layer}
		\label{fig:coupling_layer}
	\end{figure}

	The building blocks of INNs are invertible \textit{coupling layers} as proposed by Dinh et al. \cite{nice,real_nvp}. In this work, we use the modified version from Ardizzone et al. \cite{lynton} as shown in Figure \ref{fig:coupling_layer}. The coupling layer takes the input $x$ and splits it into $x_1$ and $x_2$. The neural networks $s_2$ and $t_2$, also called \textit{coupling functions} in the setting of coupling layers, take $x_2$ as input and scale/translate $x_1$ by their outputs. Afterwards, $x_2$ will be scaled and translated by the same approach. The output $y$ will be the concatenation of $y_1$ and $y_2$. Mathematically, the forward process can be described as:
	
	\begin{align}
		\label{eg:forward}
		y_1 &= x_1 \odot exp(s_2(x_2)) + t_2(x_2) \\
		y_2 &= x_2 \odot exp(s_1(y_1)) + t_1(y_1)
	\end{align} 
	
	where $s_i$ (\textit{scale}) and $t_i$ (\textit{translation}) are arbitrarily complicated coupling functions represented by classical neural networks and $\odot$ is the Hadamard product. In practice, we use the exponential function and clip extreme values to avoid numerical problems. The coupling layer is fully invertible, meaning the input $x = [x_1, x_2]$ can be reconstructed by the output $y = [y_1, y_2]$ with:
	
	\begin{align}
		\label{eg:backward}
		x_1 &= ( y_1 - t_2(x_2) ) \odot exp(-s_2(x_2)) \\
		x_2 &= ( y_2 - t_1(y_1) ) \odot exp(-s_1(y_1)) 
	\end{align}
	
\begin{figure}
		\centering
		\includegraphics[width=0.7\linewidth]{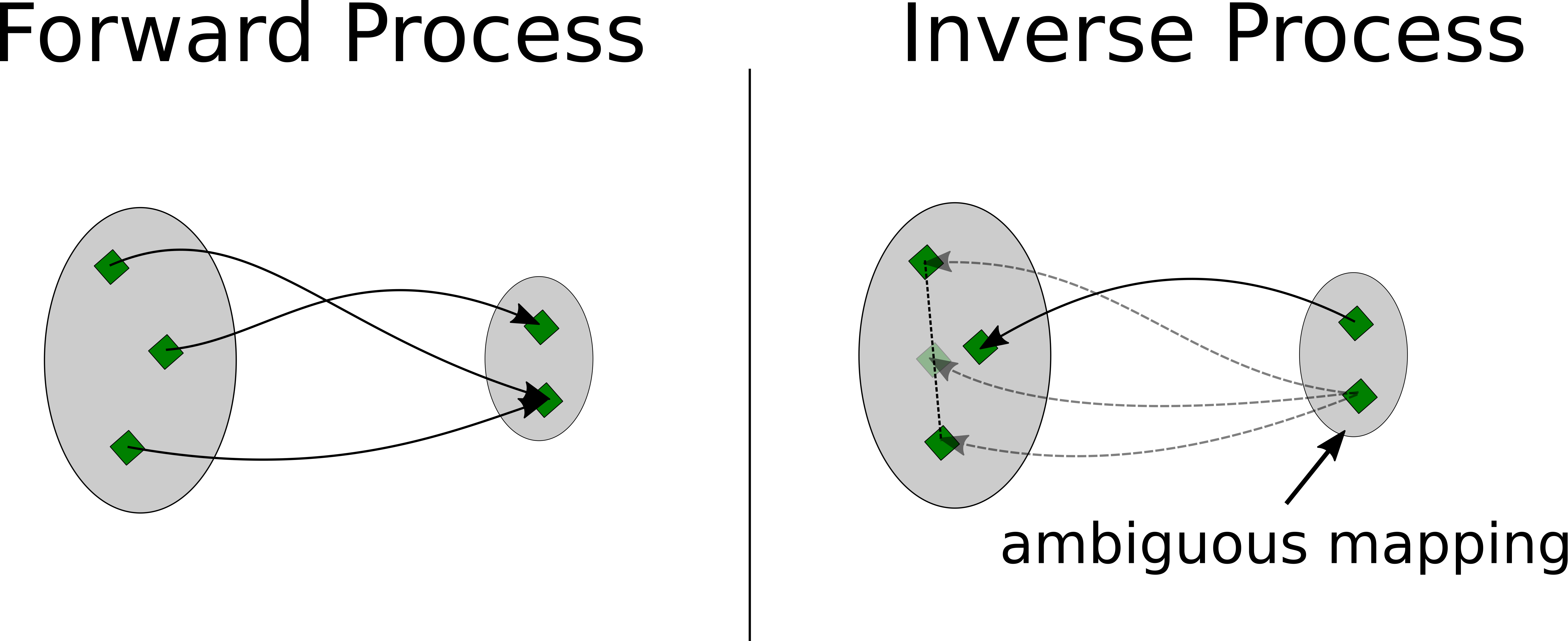} 
		\caption{Visualization of Information Loss during the Forward Process and Ambiguity of the Inverse Problem}
		\label{fig:inn_ambiguity}
	\end{figure}

	By stacking those coupling layers, we obtain an INN. In contrast to deep neural networks (DNN), which can learn any function, an INN will always learn a bijective mapping. Generally, DNNs learn a non-bijective mapping causing an inherent \textit{information loss} during the forward process, which makes the \textit{inverse process} ambiguous (see Figure \ref{fig:inn_ambiguity}). As presented in \cite{lynton}, this problem can be solved by adding latent output variables $z$ in addition to $y$. The INN is then trained to put the information lost during the forward process $x \rightarrow y$ into the latent variables $z$. In other words, the latent variables $z$ contain all the information that is not contained in $y$, but was originally part of the input $x$. As a result, the INN will learn a bijective mapping $x \leftrightarrow [y, z]$ (see Figure \ref{fig:inn_vae}).
	
	INNs allow for \textit{bi-directional training} (see \cite{lynton}). For every training iteration, the forward as well as the inverse process will be performed. This enables us to compute the gradients in both directions and optimize losses on both the input and output domains with every iteration. Bi-directional and cyclic training have improved performance of GANs \cite{gan,cgan}  and autoencoders as demonstrated by \cite{cycle_3,cycle_2,cycle_4,cycle_1}. 

\subsection{Artificial Bottleneck and INN Autoencoder}

    For INN mappings to be bijective, the dimensions of inputs $x$ and outputs $[y, z]$ have to be identical. In contrast, the classical autoencoder has a bottleneck, which allows useful representations to be learned. Without this bottleneck restriction, learning to reconstruct an image would be a trivial task. 
	
\begin{figure}
		\centering
		\includegraphics[width=0.9\linewidth]{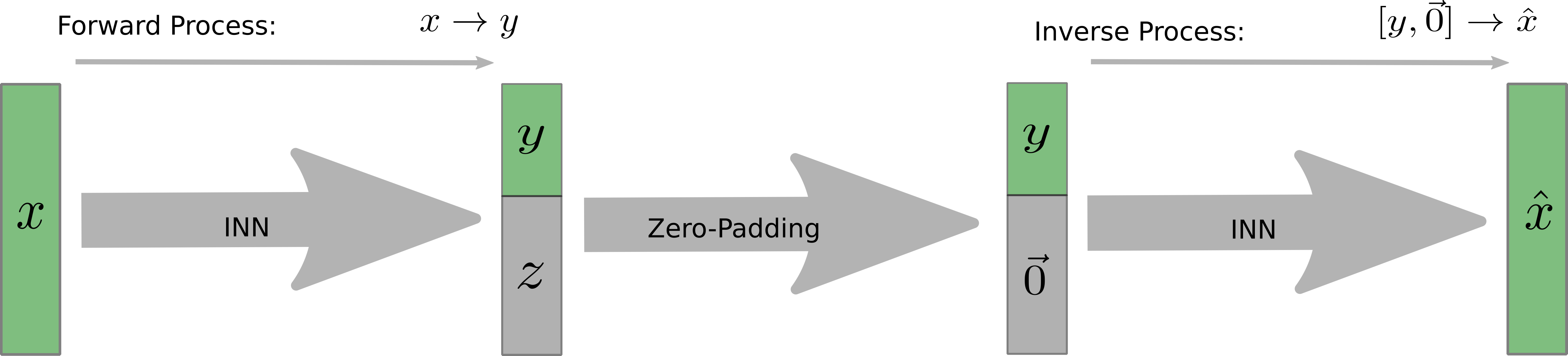}
		\caption{Visualization of Zero-Padding: Zero-Padding will be applied at all times. This ensures that the information in $z$ is not available, hence creating an artificial bottleneck.}
		\label{fig:inn_zero}
	\end{figure}

	In order to build an INN autoencoder, we need to introduce an artificial bottleneck, that emulates its classical counterpart. This is achieved by zero-padding $z$ at all times. With zero-padding, we ensure the extra dimensions given by $z$ can not be used for representation learning. As a result, the forward process of INN autoencoders is given by $x \rightarrow [y, z \neq 0]$ and the inverse by $[y, z=0] \rightarrow \hat{x}$ (see Figure \ref{fig:inn_zero}). The length of $y$ defines the bottleneck dimension of INN autoencoders. The forward process of the INN can then be interpreted as the encoder and the inverse process as the decoder.

	We train INNs as autoencoders by combining the artificial bottleneck with a reconstruction loss $L(x, \hat{x})$. The artificial bottleneck and the reconstruction loss will enforce the INN to put as much information as possible into $y$ and reduce the information contained in $z$. The information loss through zero-padding is minimized, if $z$ is the zero vector in the first place. Therefore, we add a zero-padding loss $\Omega(z, \vec{0})$ which compares $z$ to the zero vector. The zero-padding loss $\Omega$ is not essential for the INN autoencoder, since the reconstruction loss $L$ alone with the artificial bottleneck will enforce $z$ to converge against the zero vector. However, we observed that the zero-padding loss $\Omega$ slightly improves convergence rate and stability during training without negatively affecting the reconstruction quality, since its objective is aligned with the reconstruction loss. We want to emphasize that the INN autoencoder can be successfully trained without using the zero-padding loss $\Omega$. We only add the zero-padding loss $\Omega$ for technical reasons, because it does not affect reconstruction quality while making training more comfortable. The total loss function for training an INN autoencoder is given by:
	
	\begin{align}
	L(x, \hat{x}) + \Omega(z, \vec{0})
	\label{eq:inn_auto_loss}
	\end{align}

	We can extend the INN autoencoder to an INN VAE by adding a distribution loss $D$ to the loss function \eqref{eq:inn_auto_loss} which compares the learned latent distribution loss $q(y)$ with the true prior distribution $p(y)$:
	
	\begin{align}
		L(x, \hat{x}) + \Omega(z, \vec{0}) + D_{MMD} (q(y) ~||~ p(y))
		\label{eq:inn_vae_loss}
	\end{align}
	
\begin{figure}
		\centering
		\includegraphics[width=0.9\linewidth]{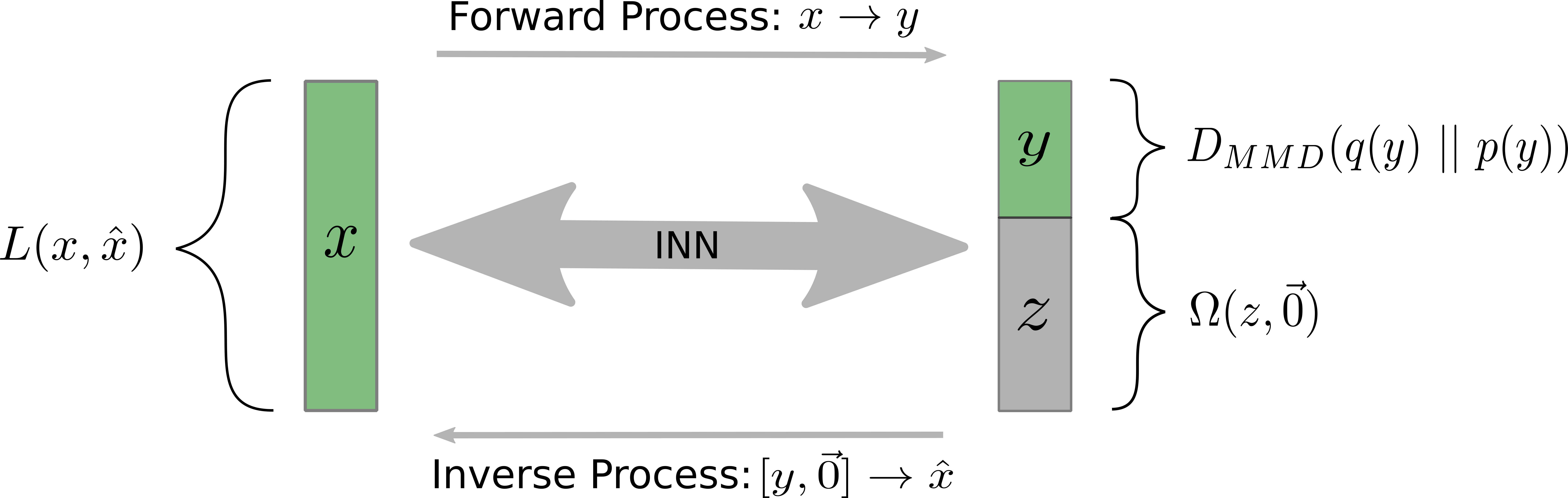}
		\caption{Visualization of Training a INN Variational Autoencoder with Reconstruction Loss $L$, Zero-Padding Loss $\Omega$ and Distribution Loss $D_{MMD}$ (For the simple INN autoencoder the distribution loss $D_{MMD}$ is omitted.)}
		\label{fig:inn_vae}
	\end{figure}

	However, instead of using KL-divergence \cite{kl_divergence} as commonly used for classical VAEs to compare distributions, we will use maximum mean discrepancy (MMD) \cite{mmd} as proposed by Ardizzone et al. \cite{lynton} to compare distributions for INNs. Figure \ref{fig:inn_vae} visualizes the loss terms in equation \eqref{eq:inn_vae_loss} for INN VAEs.
	
	\section{Experimental Setup}
	
	The goal of our experiments is to examine how our proposed INN autoencoder trained with the artificial bottleneck performs in comparison to its classical counterpart with similar number of trainable weights. We trained our models until convergence for different bottleneck sizes and accessed the reconstruction loss on the testset. This will indicate how well our models have learned the representation of a given dataset.
	
	\subsection{Architecture}
	
	For MNIST, we trained four different classical models. The encoder of model \textit{classic} consists of four fully-connected layers (hidden size: 512, 256, 128 and bottleneck size) followed by ReLUs. The encoder of model \textit{classic 1024} and \textit{classic 2048} follow the same architecture, however the hidden size are modified to be 1024, 1024, 1024, bottleneck size and 2048, 2048, 2048, bottleneck size respectively. The model \textit{classic deep} follows the same architecture as model \textit{classic 1024}, however it has two additional fully-connected layers of size 1024. For CIFAR-10 and CelebA, the encoder consists of five convolutional layers (CIFAR-10: kernel size 3, stride 1 / CelebA: kernel size 4, stride 2) and one fully-connected layer. For all models the decoder mirrors the encoder, whereby the last activation function is replaced by a tanh function. 
	
	The INN autoencoder on MNIST consists of three coupling layers with convolutional coupling functions (hidden channel size: 100, kernel size: 3 and leaky ReLU slope: 0.1) and one coupling layer with fully-connected coupling functions (hidden layer size: 180). On CIFAR-10, we use the same architecture as for MNIST. However, the convolutional coupling functions have hidden channel size 128 and the fully-connected coupling functions have hidden layer size 1000. For CelebA, our INN model consists of six convolutional coupling layers with same coupling functions as for CIFAR-10 and one fully-connected coupling layer with hidden layer size 200.
	
	\subsection{Training}
	
	For both classical and INN models, we used L1-norm as reconstruction loss. The zero-padding loss $\Omega$ is chosen to be L2-norm. All models are trained with adaptive learning using Adam optimization \cite{adam} (weight decay $\rightarrow$ classic: $1 \times 10^{-5}$, INN: $1 \times 10^{-6}$). 
	
	The classical models were trained for (MNIST/CIFAR-10: 100, CelebA: 10) epochs with a batch size of (MNIST/CIFAR-10: 128, CelebA: 32). The learning rate started at $1 \times 10^{-3}$ and was decreased by a factor of 10 at (MNIST: every 10th, CIFAR-10: 60th and 85th, CelebA: 8th and 9th) training epoch. 
	
	The INN models was trained for (MNIST: 10, CIFAR-10: 15, CelebA: 8) epochs with batch size (MNIST/CIFAR-10: 128, CelebA: 32). The learning rate started at $1 \times 10^{-3}$ and was decreased by a factor of 10 at the (MNIST: 8th, CIFAR-10: 10th, CelebA: 6th and 7th) epoch.

	\section{Results and Discussion}
	
\begin{figure}
		\centering
		\subfigure[Reconstruction MNIST]{\includegraphics[width=0.30\linewidth]{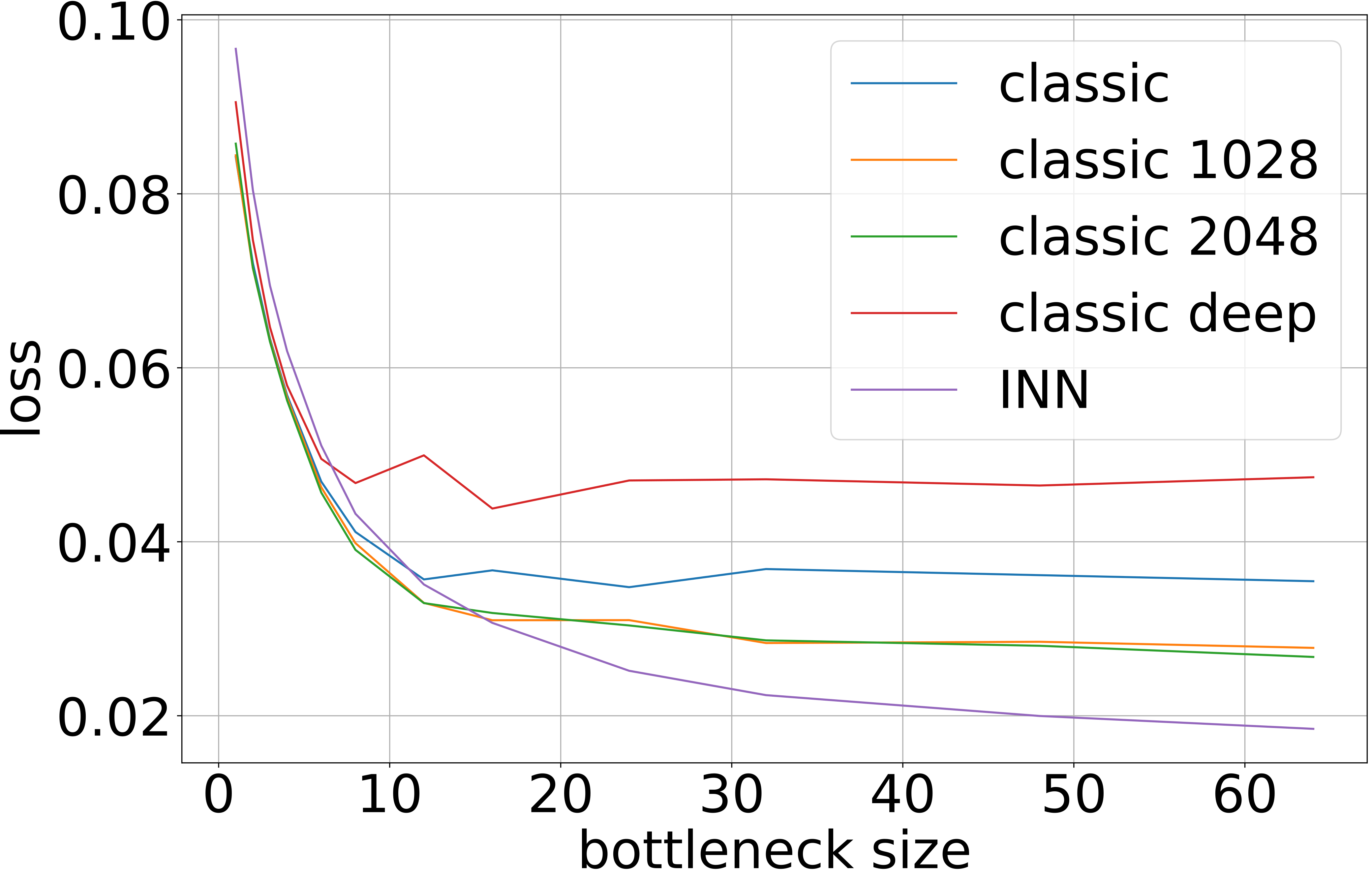}}
		\subfigure[Parameter MNIST]{\includegraphics[width=0.30\linewidth]{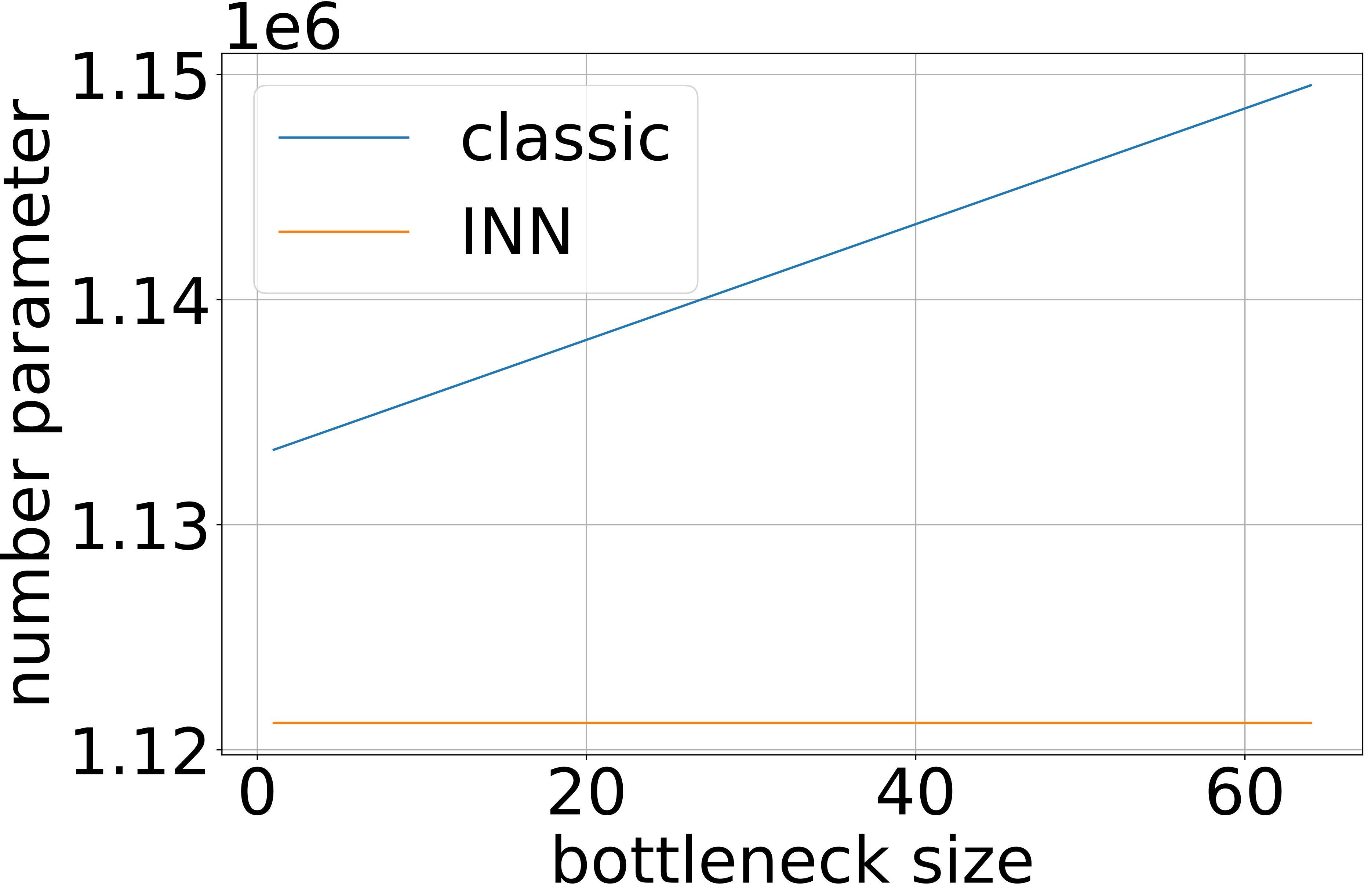}}
		\subfigure[Reconstruction CIFAR-10]{\includegraphics[width=0.30\linewidth]{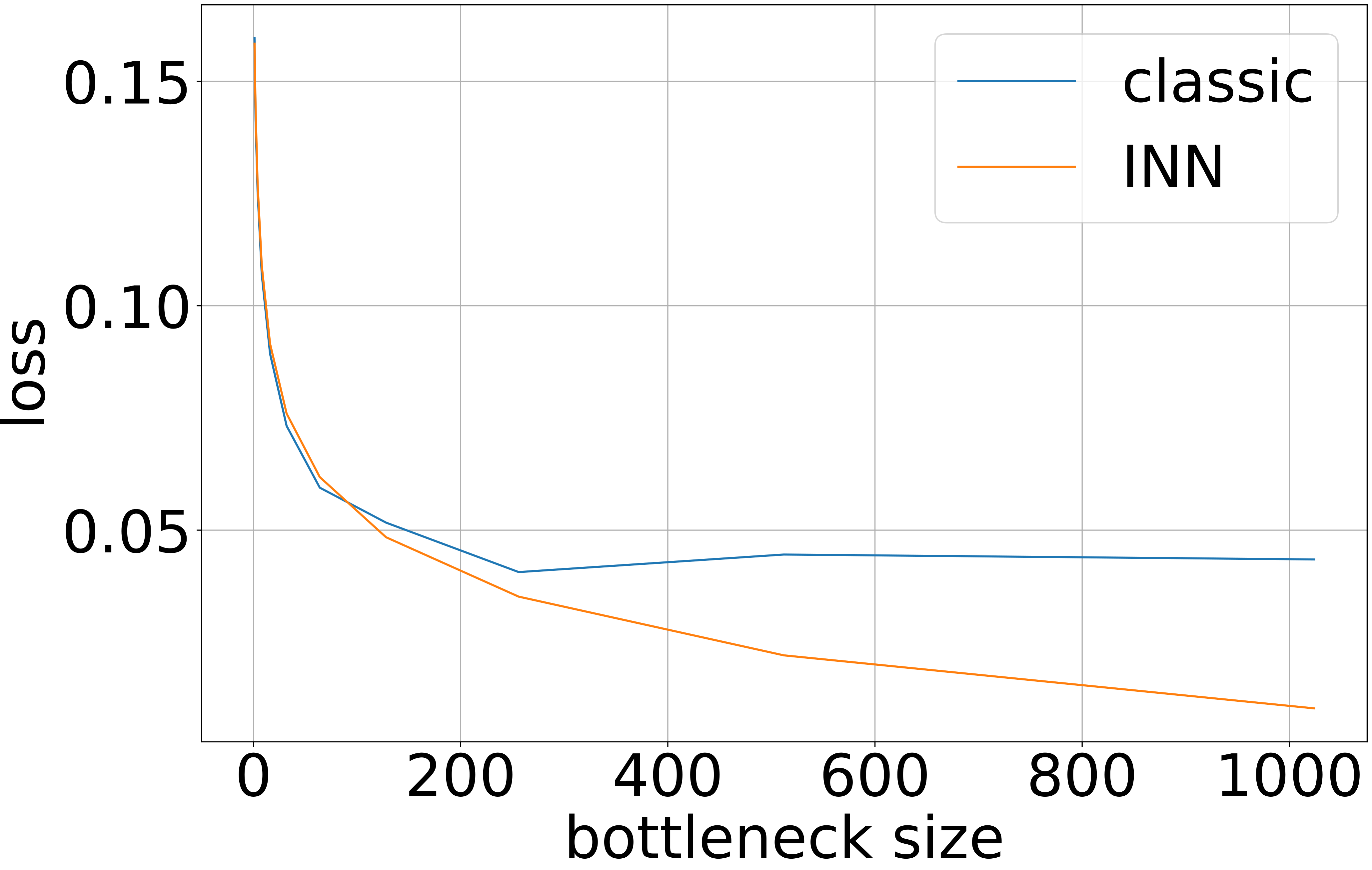}}
		\subfigure[Parameter CIFAR-10]{\includegraphics[width=0.30\linewidth]{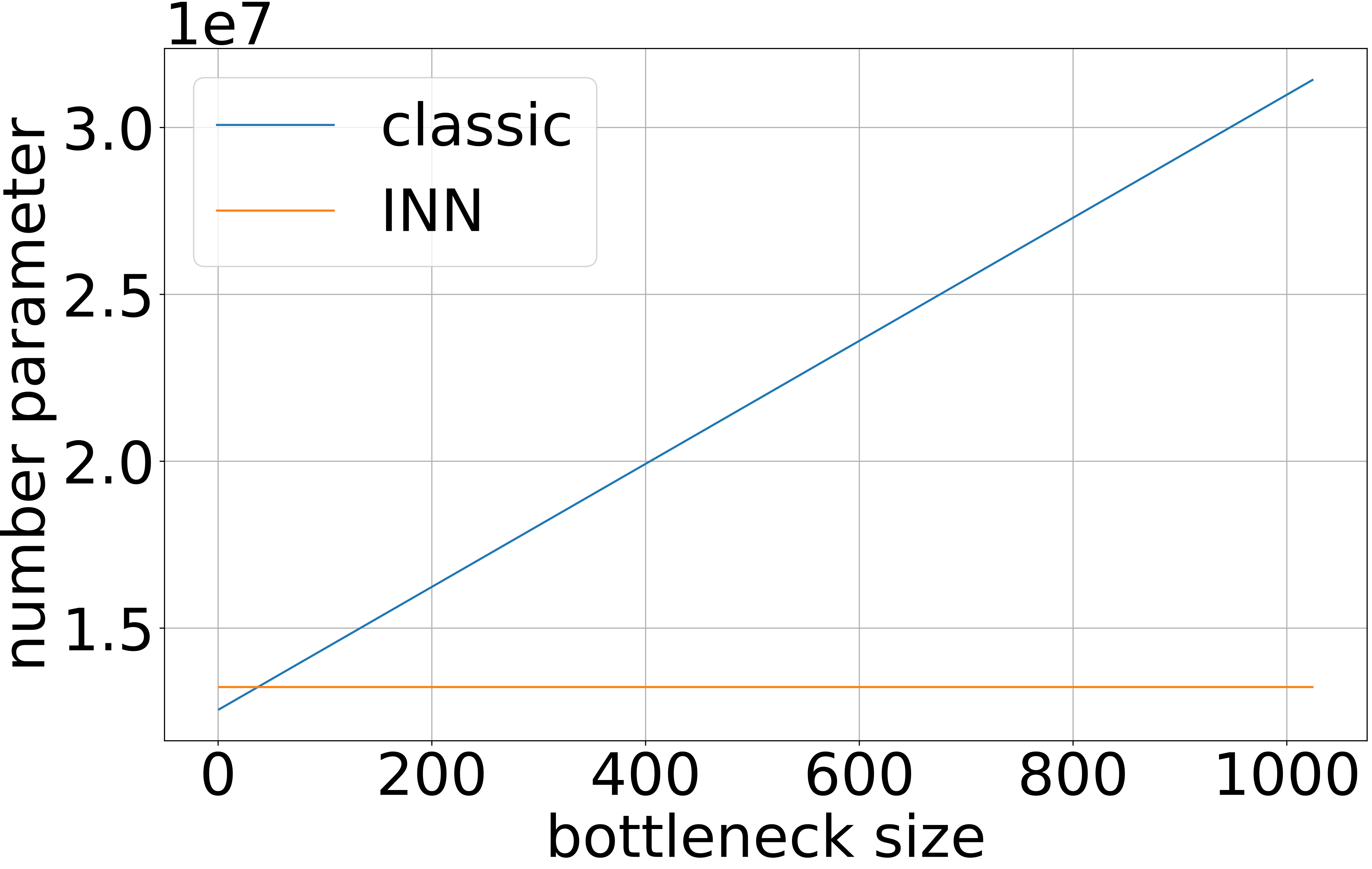}}
		\subfigure[Reconstruction CelebA]{\includegraphics[width=0.30\linewidth]{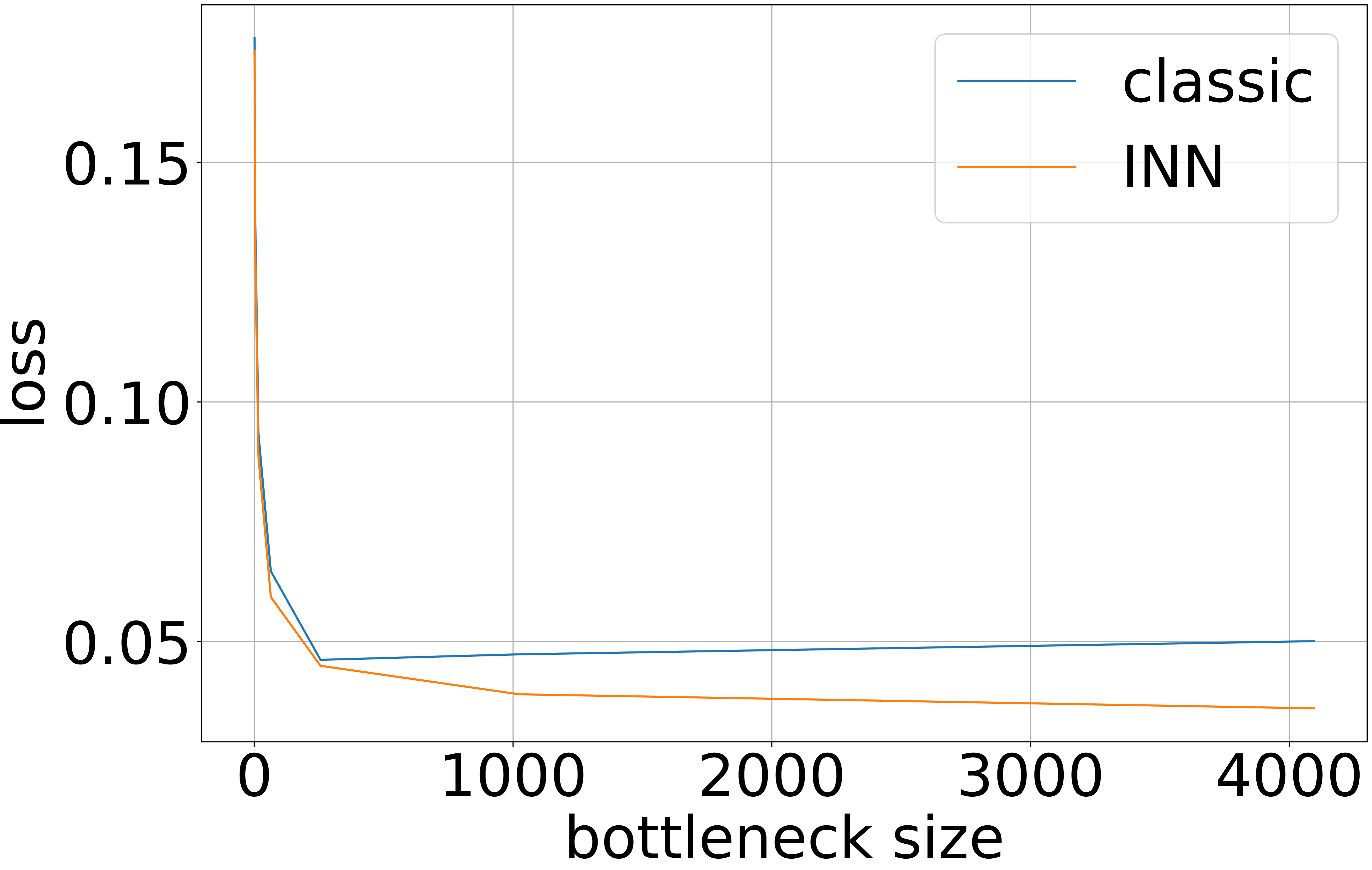}}
		\subfigure[Parameter CelebA]{\includegraphics[width=0.30\linewidth]{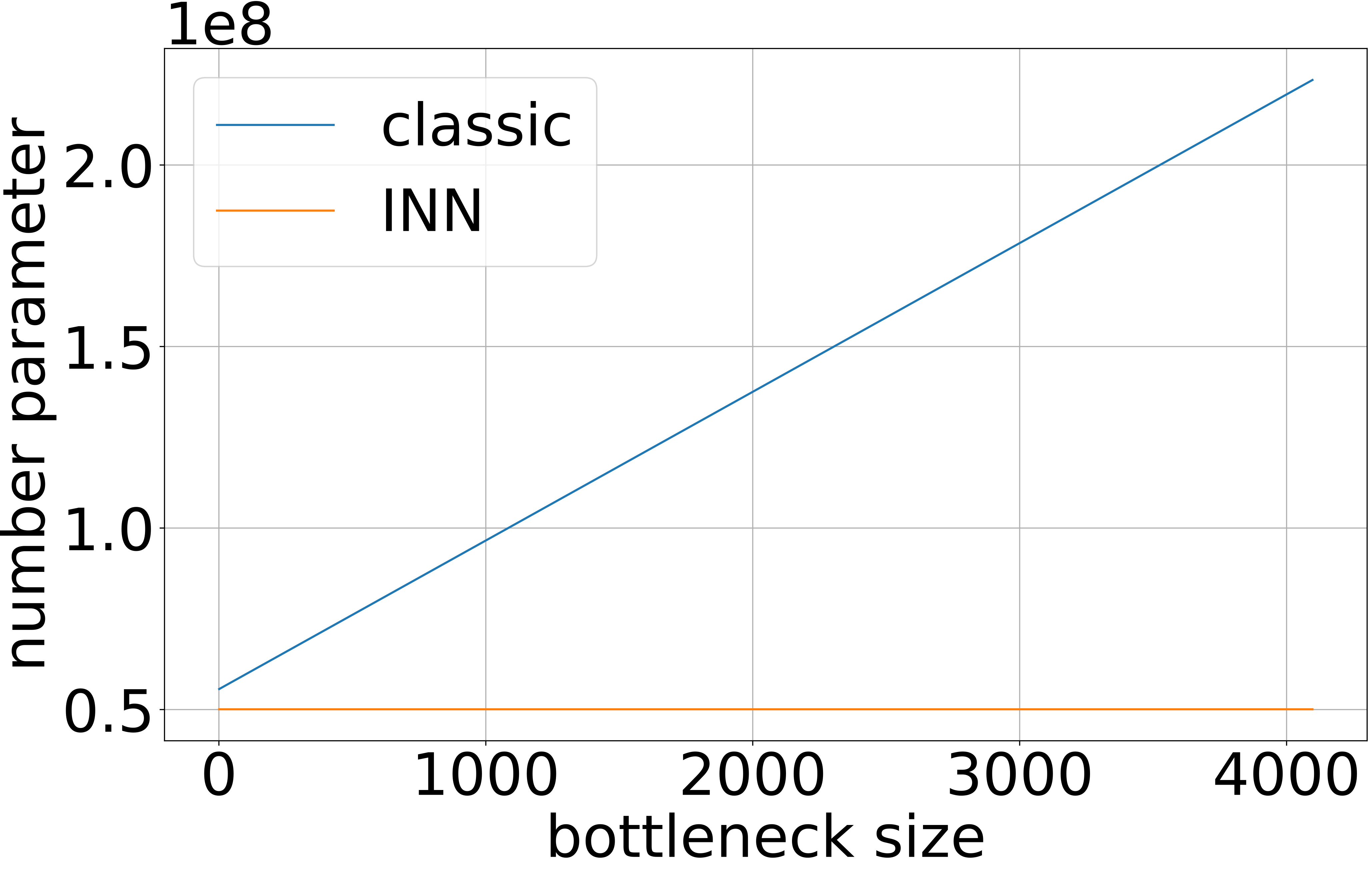}}
		\caption{Comparison of Test Reconstruction Loss between our Classical and INN Models (a, c, e) and Corresponding Number of Trainable Parameters (b, d, f) for Different Bottleneck Sizes on MNIST, CIFAR-10 and CelebA. Note that the y-axis of b, d and f does not necessary start at zero.}
		\label{fig:result}
	\end{figure}

	In Figure \ref{fig:result}, we summarized our results by plotting the test reconstruction loss and the corresponding number of parameters in our models against the bottleneck size for MNIST, CIFAR-10 and CelebA.   
	
	We observe that across all three datasets, the INN autoencoder delivers results comparable to its classical counterpart for small bottleneck sizes. 
	
	Initially, we expected the reconstruction loss to further decrease for larger bottleneck sizes for both classical and INN models. However, the reconstruction loss for classical autoencoders seems to saturate for larger bottleneck sizes showing no further improvement despite increasing the bottleneck (see Figure \ref{fig:result}). The saturation sets in at about bottleneck sizes of 12 (MNIST), 250 (CIFAR-10) and 200-250 (CelebA) \footnote{Due to hardware/GPU limitations, we only trained our CelebA models for a sparse number of bottleneck sizes. This makes it more difficult to determine the exact point at which saturation sets in}. Based on the results, it seems, that saturation sets in at the approximate intrinsic dimension of the datasets.
	
\begin{figure}
		\centering
		\subfigure[Input (INN)]{\includegraphics[width=0.40\textwidth]{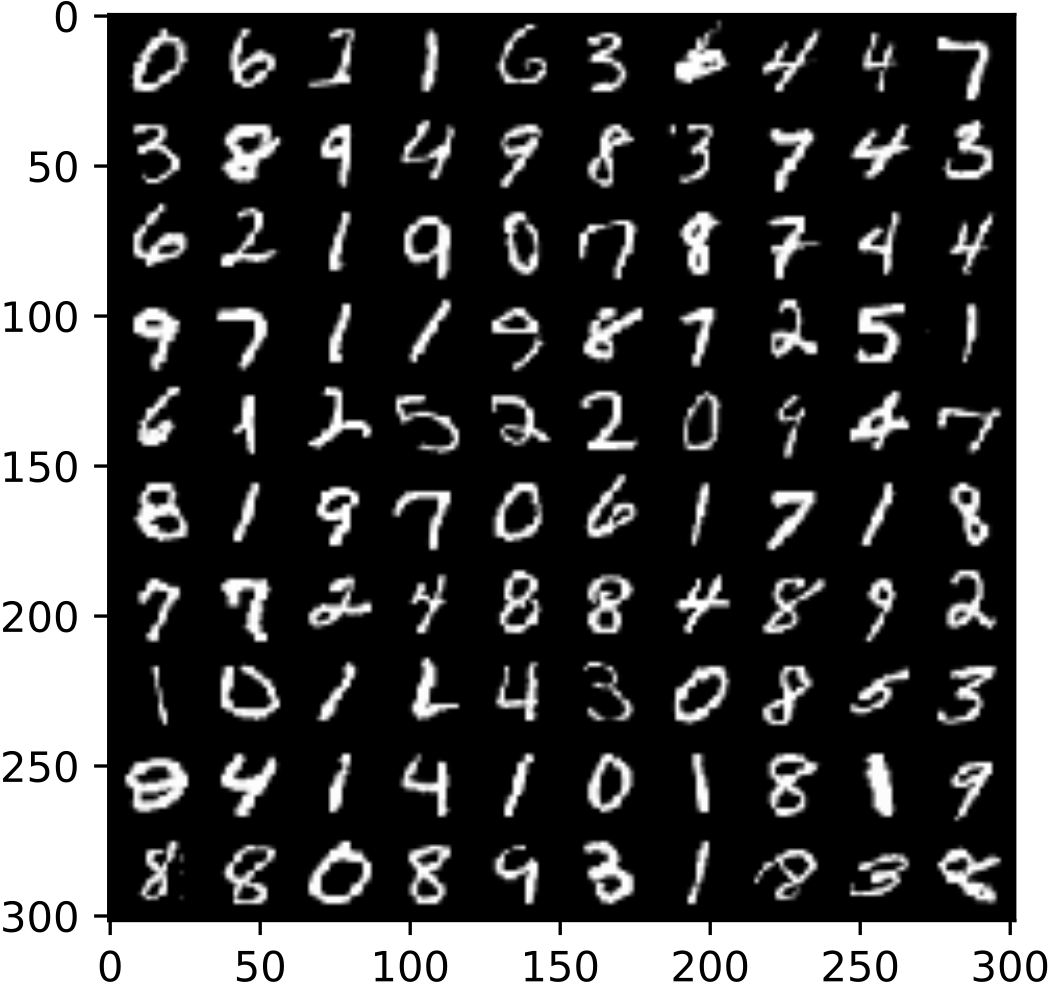}}
		\subfigure[Input (classic)]{\includegraphics[width=0.40\textwidth]{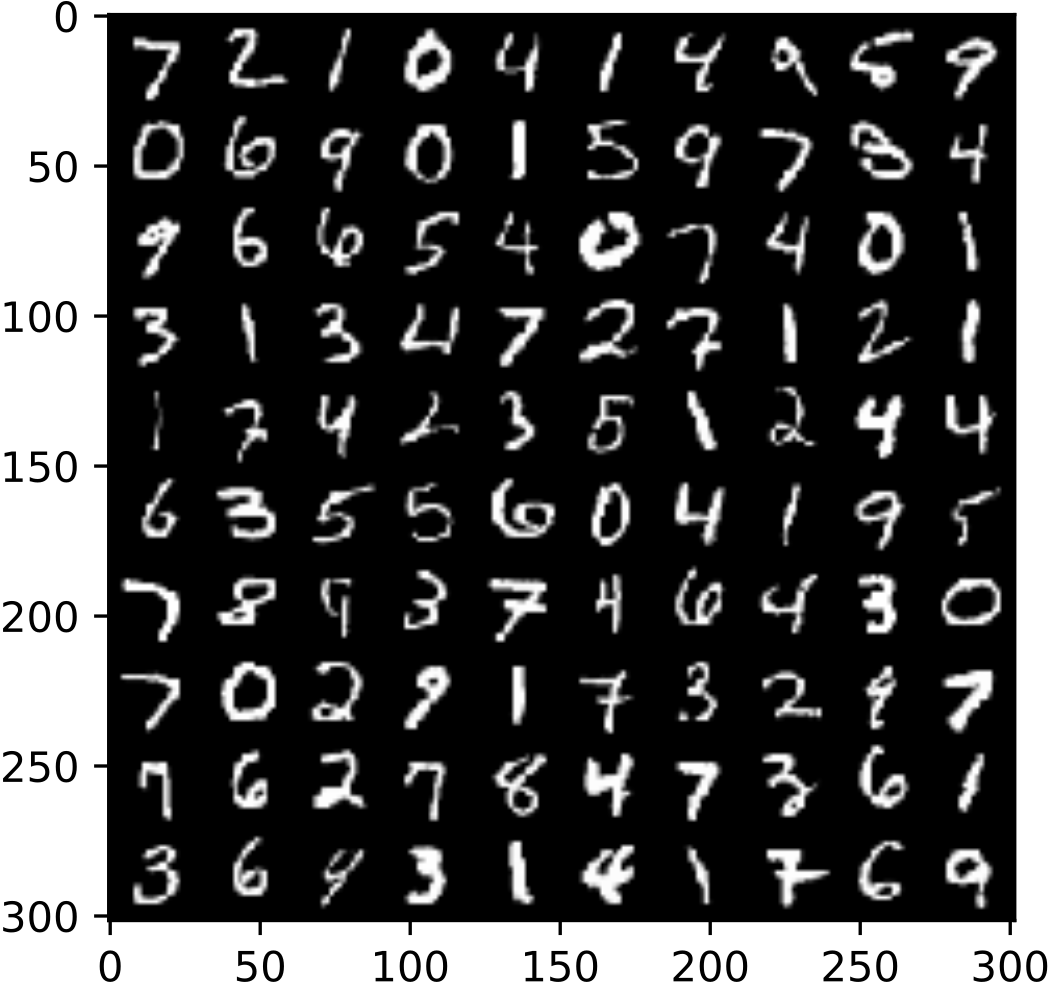}}
		\subfigure[Reconstruction (INN)]{\includegraphics[width=0.40\textwidth]{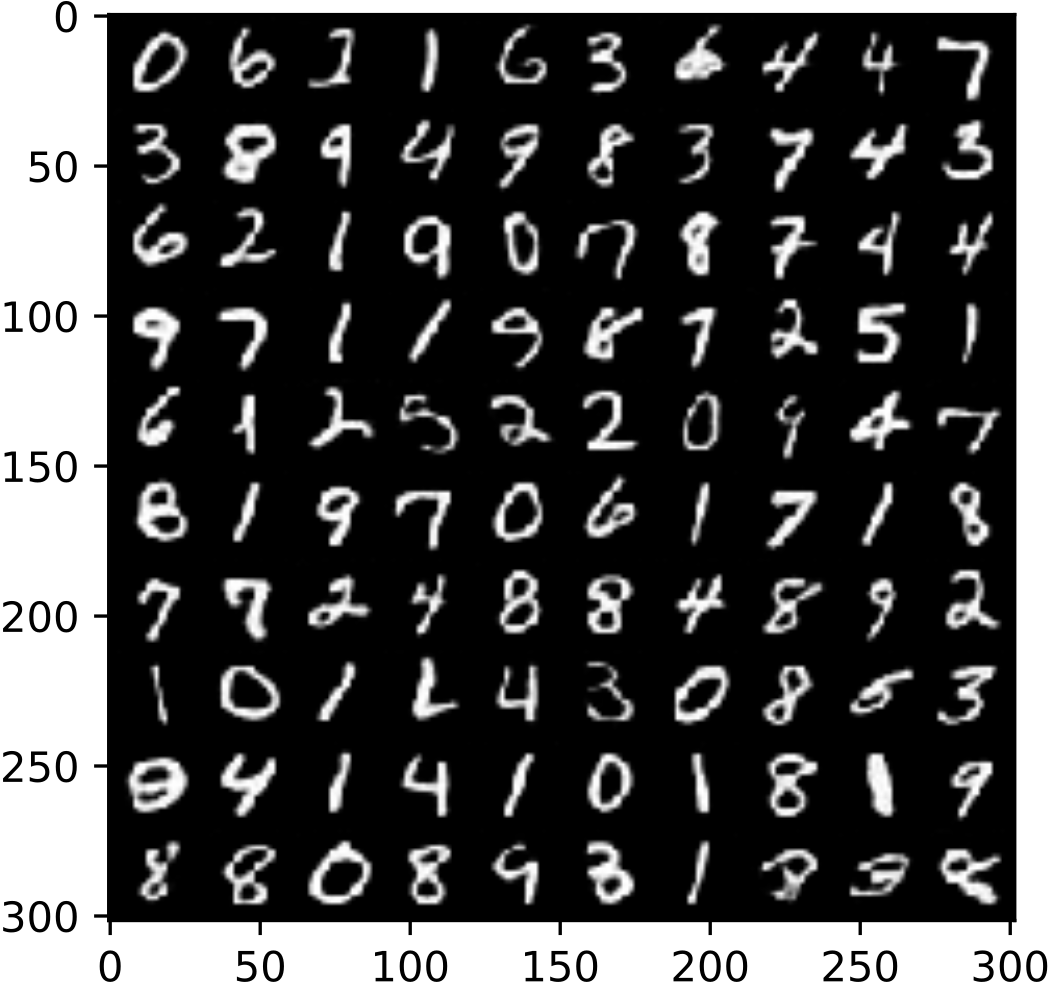}}
		\subfigure[Reconstruction (classic)]{\includegraphics[width=0.40\textwidth]{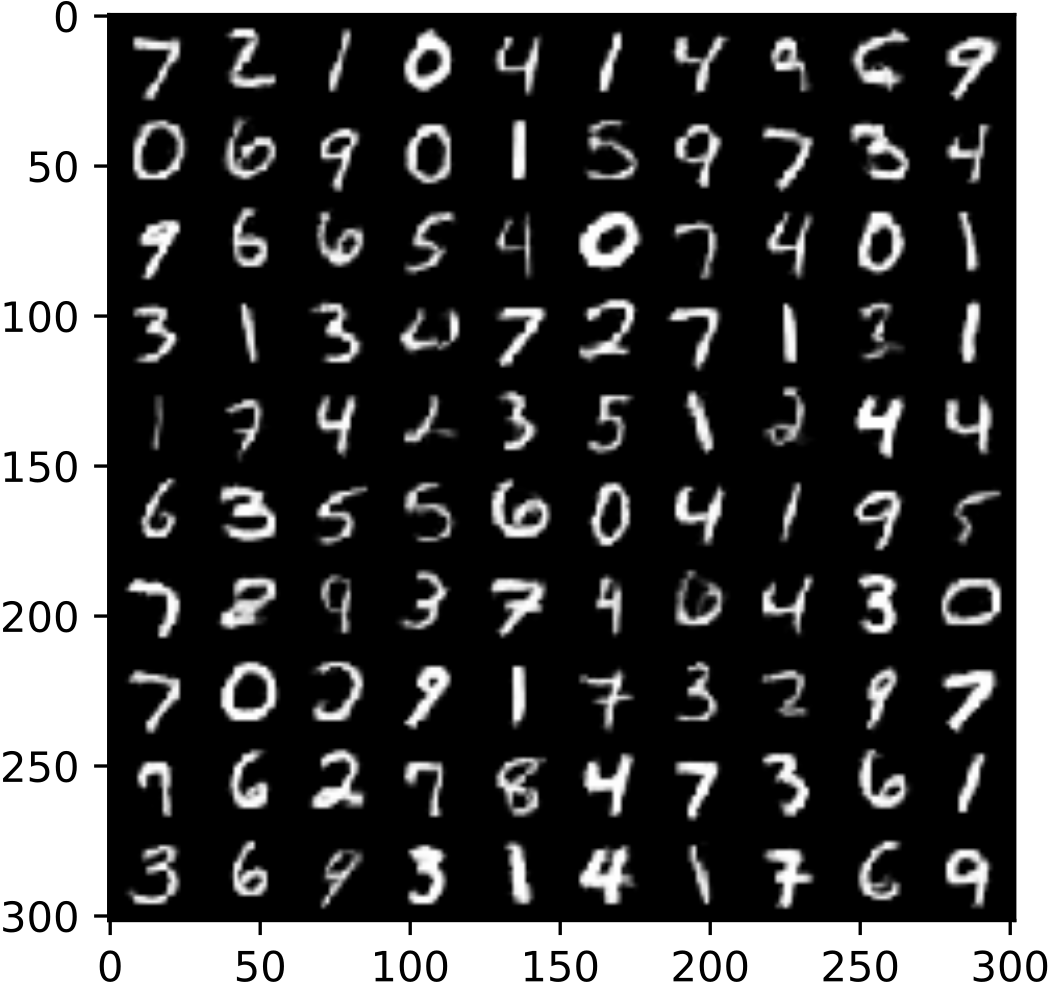}}
		\subfigure[Difference (INN)]{\includegraphics[width=0.40\textwidth]{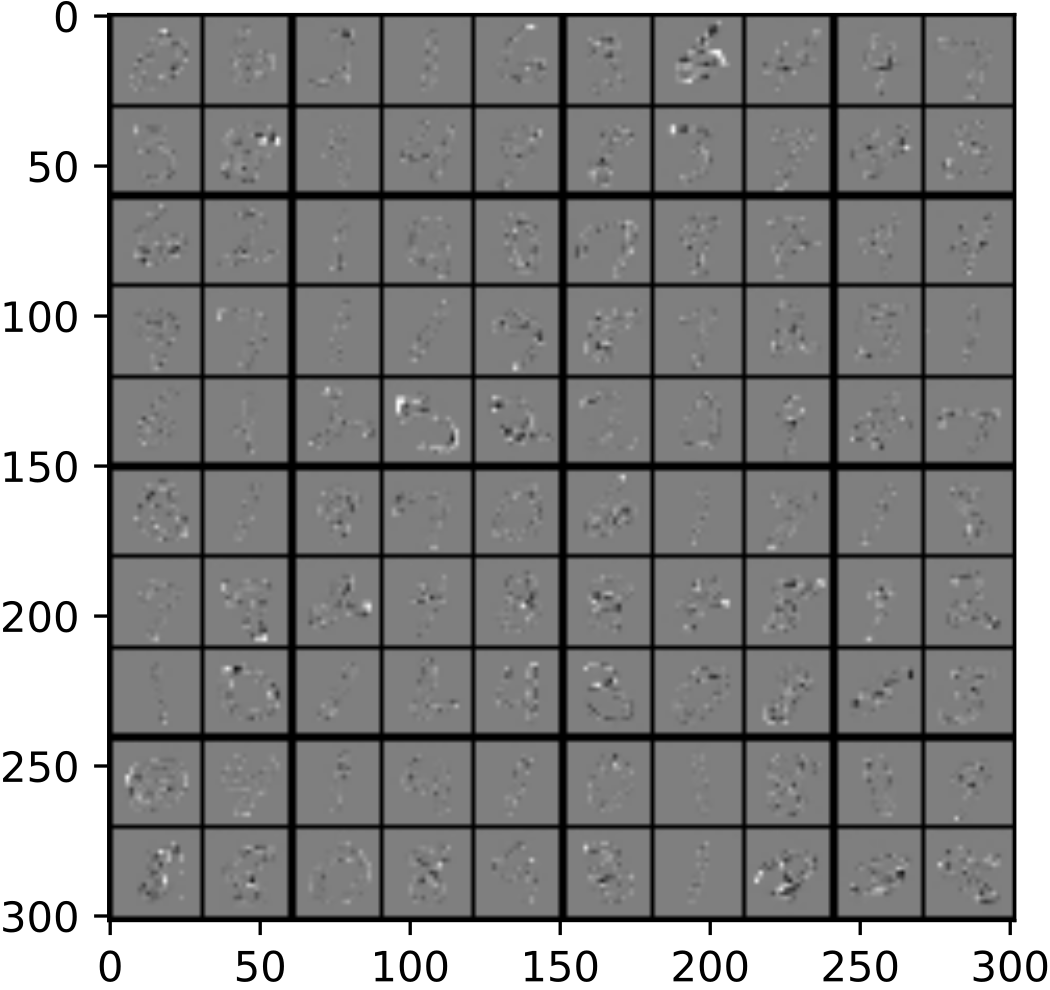}}
		\subfigure[Difference (classic)]{\includegraphics[width=0.40\textwidth]{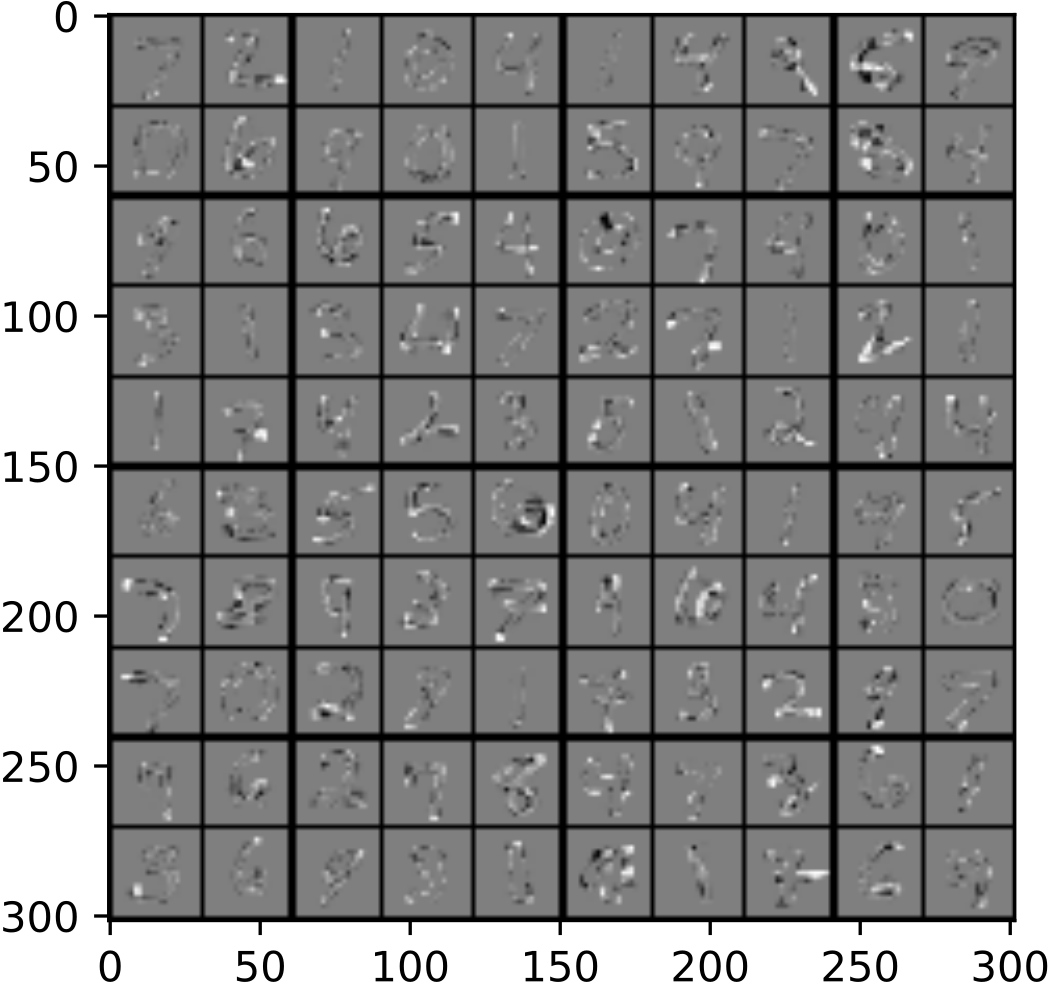}}
		\caption{Example of Random Reconstructed MNIST Images by our INN (left) and Classical (right) Autoencoder with Bottleneck Size 32: \textit{Input} shows the original input images from the MNIST testset, \textit{Reconstruction} shows the reconstructed images, \textit{Difference} shows the difference between original and reconstructed image.}
		\label{fig:res_example}
	\end{figure}

	In contrast, the INN autoencoder reconstruction loss resembles the expected exponential decay curve with better performance for larger bottleneck sizes. Since the INN autoencoder reconstruction loss does not saturate, it performs significantly better for larger bottleneck sizes than the classical autoencoder. In Figure \ref{fig:res_example}, we show examples of randomly reconstructed MNIST images by our INN and classical model with bottleneck size 32. Especially the difference between original and reconstructed image shows that the INN autoencoder produces better reconstructions than the classical autoencoder.   
	
	Besides the reconstruction loss, we additionally compared the number of trainable parameters in our models for different bottleneck sizes (see Figure \ref{fig:result}). The INN autoencoder architecture does not have to be changed for varying bottleneck sizes, only the split between $y$ and $z$ needs to be redefined. For the classical autoencoder, at least the bottleneck layer has to be changed, resulting in a higher number of trainable parameters for larger bottleneck sizes.
	
	Since the saturation in reconstruction loss for our classical models was quite unexpected, we conducted additional experiments to further investigate the cause of observed saturation. For MNIST, we trained three additional models: \textit{classic 1024}, \textit{classic 2048} and \textit{classic deep}. 
	
	Even though the reconstruction loss of model \textit{classic 1024} and \textit{classic 2048} improves for larger bottleneck sizes compared to model \textit{classic}, it still does not reach the same performance as our INN model. The difference in performance between model \textit{classic 1024} and \textit{classic 2048} is relatively small taking into account that model \textit{classic 2048} has twice the hidden layer size of model \textit{classic 1024}. This indicates that further increasing the hidden layer size would not yield significant improvements. 
	
	Model \textit{classic deep} saturates at an even higher reconstruction loss than all other models, despite being deeper. This lets us conclude that model \textit{classic} was already deep and complex enough for the given task on MNIST. This eliminates the possibility that our model \textit{classic} fails to achieve the same reconstruction performance as our INN model simply because it was not deep enough. 
	
	Despite all the models being different, the saturation still sets in at about the same bottleneck size of 12 for all four classical models (see Figure \ref{fig:result}).
	
	It is important to note that the hyperparameters for our models were only optimized for bottleneck sizes of 12 (MNIST) and 300 (CIFAR-10, CelebA). For all other bottleneck sizes, we trained our models with the same hyperparameters. However, for our MNIST classical models, we additionally optimized the hyperparameters for bottleneck size 64. Even with optimal hyperparameters, we could not reach the same performance as with our INN model. We also checked the reconstruction loss of our classical models on both the train- and testset and found no significant divergence which rules out the possibility of overfitting.
	
	Given the results outlined above, we hypothesize that the cause of observed saturation leads back to the fundamental difference in network architecture of classical and INN autoencoders. Our hypothesis builds upon recent works of Yu et al. They applied information theory to CNNs \cite{info_loss_2} and specifically to autoencoders in \cite{info_loss_1}. By measuring the mutual information between input and feature maps of various layers within a Deep Neural Network (DNN), they found that the mutual information decreases for deeper layers. Therefore, they concluded: \textit{"However, from the DPI perspective validated in this work ([...]), the deeper the neural networks, the more information about the input is lost, thus the less information the network can manipulate. In this sense, one can expect an upper bound on the number of layers in DNNs that achieves optimal performance."} (cited from \cite{info_loss_1}). We believe that this is exactly what we observe with our model \textit{classic deep}. It would also explain why the model \textit{classic deep} performs worse than the other three models. Due to its depth, it loses more information about the input than all the other classical models. Further works that take an information-theoretic view and investigated the information bottleneck are \cite{info_bottle_3,info_bottle_2,info_bottle_1}. 
	
	This leads us back to the ambiguity problem of DNNs. We already established, that if a DNN does not learn a bijective function, information loss occurs during the forward process making the inverse process ambiguous. The INN solves this ambiguity problem by introducing latent variables $z$ containing all the information lost during the forward process (see \cite{lynton}). Therefore, we hypothesize that INNs have no intrinsic information loss contrary to DNNs and the findings of Yu et al. \cite{info_loss_1} and \cite{info_loss_2} do not apply to INNs. In other words, INNs are not bound to a maximal number of layers (depth) after which only suboptimal results can be achieved. 
	
	Furthermore, we believe that the intrinsic information loss of DNNs causes the saturation. In reverse, we could interpret the threshold at which the reconstruction loss saturates as a quantification for the intrinsic information loss of our classical models. This would explain why our deepest model \textit{classic deep} saturates at the highest reconstruction loss threshold, since it has the highest intrinsic information loss. 
	
	Increasing the hidden layer size as done with model \textit{classic 1024} and \textit{classic 2048}, seems to reduce the intrinsic information loss of the architecture. This makes intuitive sense, since a fully-connected layer of larger size is able to extract more information and minimize the information loss between its input and output. However, our experiments suggest that there is a lower bound for the information loss of DNNs at which increasing the hidden layer size will not further decrease its intrinsic information loss. 
	
	Since INN autoencoders do not have any intrinsic information loss according to our hypothesis, we expect the reconstruction loss threshold \footnote{at which saturation occurs} of INN autoencoders to be at zero. 
	
	Despite INNs having no intrinsic information loss and thereby allowing arbitrary deep designs, in the application of autoencoders this might be a disadvantage. The main idea of autoencoders is to extract the essential information of the dataset and discard the rest to achieve a dimensionality reduction. Bounded to an information loss, a DNN has to get rid of information within its input. Most likely, it will choose to remove the noise in the dataset and keep the most essential information. This explains why the saturation sets in at approximately the intrinsic dimension of the datasets. At this bottleneck size the whole essential information of the dataset is already encoded. Further increasing the bottleneck would not yield any improvement, since the noise was already removed during the forward process. In contrast, the INN keeps all the noise within the latent variables $z$. Further increasing the bottleneck would add noise to the $y$-latent space in addition to the essential information and further improve the reconstruction quality. In case of dimensionality reduction, this might be an undesirable characteristic, since the goal is to remove the noise. 
	
	We conclude, that in order to use INN autoencoders for dimensionality reduction, the intrinsic dimension of the dataset has to be known or estimate appropriately. Nevertheless, if the bottleneck size is chosen accordingly, the INN autoencoder is capable of just learning the essential information of the dataset, but at the same time leaves the option to additionally learn the noise if needed. Furthermore, the INN autoencoder preserve all the advantages of INNs.

	\section{Conclusion}
	
	In conclusion, the experiments show that our proposed method of training INNs as autoencoders does indeed work. For small bottleneck sizes, the INN autoencoder performs equally well as the classical autoencoder. It is capable of extracting the essential information of a dataset and learning useful representations. For large bottleneck sizes, greater than the intrinsic dimension of the dataset, the INN autoencoder outperforms its classical counterpart. In summary, we demonstrated that the architecture restrictions caused by the invertibility constraint does not negatively affect the performance, while all the advantages of INNs such as tractable Jacobian for both forward and inverse mapping as well as explicit computation of posterior probabilities are still preserved. We hypothesize that INNs do not have any intrinsic information loss. This would allow the INN autoencoder to additionally learn the noise of the dataset, if the bottleneck size is chosen larger than the intrinsic dimension of the dataset. Furthermore, INNs might not be bounded to a maximal number of layers (depth) after which only suboptimal results can be achieved. However, further research is necessary to validate these hypotheses. Another advantage of INN autoencoders is that they are more versatile across different bottleneck sizes. The architecture does not need to be changed, only the split between $y$ and $z$ has to be redefined. There are both advantages as well as disadvantages in using INN autoencoders compared to the classical approach which we have outlined in this work. However, we believe that our results have shown interesting properties of INNs and their innate differences from classical DNNs. At the moment, INNs are still not fully understood which leaves room for further research. We hope that our findings can help towards uncovering the properties of INNs and encourage further research in this direction. 

    \section{Acknowledgment}
    This work is supported by the Bundesministerium fuer Wirtschaft und Klimaschutz (BMWK, German Federal Ministry for Economic Affairs and Climate Action) as part of the German government’s 7th energy research program "Innovations for the energy transition" under the 03ETE039I HiBRAIN project (Holistic method of a combined data- and model-based Electrode design supported by artificial intelligence) and the  Deutsche Forschungsgemeinschaft (DFG, German Research Foundation) under Germany’s Excellence Strategy EXC-2181/1 - 390900948 Heidelberg STRUCTURES Cluster of Excellence.

\bibliographystyle{unsrt}  
\bibliography{references}

\begin{thebibliography}{10}

\bibitem{object_detection}
Ross {Girshick}.
\newblock {Fast R-CNN}.
\newblock {\em arXiv e-prints}, page arXiv:1504.08083, Apr 2015.

\bibitem{image_captioning}
Andrej {Karpathy} and Li~{Fei-Fei}.
\newblock {Deep Visual-Semantic Alignments for Generating Image Descriptions}.
\newblock {\em arXiv e-prints}, page arXiv:1412.2306, Dec 2014.

\bibitem{semantic_segmentation}
Jonathan {Long}, Evan {Shelhamer}, and Trevor {Darrell}.
\newblock {Fully Convolutional Networks for Semantic Segmentation}.
\newblock {\em arXiv e-prints}, page arXiv:1411.4038, Nov 2014.

\bibitem{object_recognition}
Olga {Russakovsky}, Jia {Deng}, Hao {Su}, Jonathan {Krause}, Sanjeev
  {Satheesh}, Sean {Ma}, Zhiheng {Huang}, Andrej {Karpathy}, Aditya {Khosla},
  Michael {Bernstein}, Alexander~C. {Berg}, and Li~{Fei-Fei}.
\newblock {ImageNet Large Scale Visual Recognition Challenge}.
\newblock {\em arXiv e-prints}, page arXiv:1409.0575, Sep 2014.

\bibitem{scene_classification}
Bolei Zhou, Agata Lapedriza, Jianxiong Xiao, Antonio Torralba, and Aude Oliva.
\newblock Learning deep features for scene recognition using places database.
\newblock In Z.~Ghahramani, M.~Welling, C.~Cortes, N.~D. Lawrence, and K.~Q.
  Weinberger, editors, {\em Advances in Neural Information Processing Systems
  27}, pages 487--495. Curran Associates, Inc., 2014.

\bibitem{variational_inference}
David~M. {Blei}, Alp {Kucukelbir}, and Jon~D. {McAuliffe}.
\newblock {Variational Inference: A Review for Statisticians}.
\newblock {\em arXiv e-prints}, page arXiv:1601.00670, Jan 2016.

\bibitem{tutorial_vae}
Carl {Doersch}.
\newblock {Tutorial on Variational Autoencoders}.
\newblock {\em arXiv e-prints}, page arXiv:1606.05908, Jun 2016.

\bibitem{vae}
Diederik~P {Kingma} and Max {Welling}.
\newblock {Auto-Encoding Variational Bayes}.
\newblock {\em arXiv e-prints}, page arXiv:1312.6114, Dec 2013.

\bibitem{nice}
Laurent {Dinh}, David {Krueger}, and Yoshua {Bengio}.
\newblock {NICE: Non-linear Independent Components Estimation}.
\newblock {\em arXiv e-prints}, page arXiv:1410.8516, Oct 2014.

\bibitem{real_nvp}
Laurent {Dinh}, Jascha {Sohl-Dickstein}, and Samy {Bengio}.
\newblock {Density estimation using Real NVP}.
\newblock {\em arXiv e-prints}, page arXiv:1605.08803, May 2016.

\bibitem{rev_residual_net}
Aidan~N. {Gomez}, Mengye {Ren}, Raquel {Urtasun}, and Roger~B. {Grosse}.
\newblock {The Reversible Residual Network: Backpropagation Without Storing
  Activations}.
\newblock {\em arXiv e-prints}, page arXiv:1707.04585, Jul 2017.

\bibitem{resnet}
Kaiming {He}, Xiangyu {Zhang}, Shaoqing {Ren}, and Jian {Sun}.
\newblock {Deep Residual Learning for Image Recognition}.
\newblock {\em arXiv e-prints}, page arXiv:1512.03385, Dec 2015.

\bibitem{i_revnet}
J{\"o}rn-Henrik {Jacobsen}, Arnold {Smeulders}, and Edouard {Oyallon}.
\newblock {i-RevNet: Deep Invertible Networks}.
\newblock {\em arXiv e-prints}, page arXiv:1802.07088, Feb 2018.

\bibitem{glow}
Diederik~P. {Kingma} and Prafulla {Dhariwal}.
\newblock {Glow: Generative Flow with Invertible 1x1 Convolutions}.
\newblock {\em arXiv e-prints}, page arXiv:1807.03039, Jul 2018.

\bibitem{gan_real_nvp}
Ivo {Danihelka}, Balaji {Lakshminarayanan}, Benigno {Uria}, Daan {Wierstra},
  and Peter {Dayan}.
\newblock {Comparison of Maximum Likelihood and GAN-based training of Real
  NVPs}.
\newblock {\em arXiv e-prints}, page arXiv:1705.05263, May 2017.

\bibitem{generative_reversible_net}
Robin {Tibor Schirrmeister}, Patryk {Chrab{\k{a}}szcz}, Frank {Hutter}, and
  Tonio {Ball}.
\newblock {Training Generative Reversible Networks}.
\newblock {\em arXiv e-prints}, page arXiv:1806.01610, Jun 2018.

\bibitem{flow_gan}
Aditya {Grover}, Manik {Dhar}, and Stefano {Ermon}.
\newblock {Flow-GAN: Combining Maximum Likelihood and Adversarial Learning in
  Generative Models}.
\newblock {\em arXiv e-prints}, page arXiv:1705.08868, May 2017.

\bibitem{lynton}
Lynton {Ardizzone}, Jakob {Kruse}, Sebastian {Wirkert}, Daniel {Rahner},
  Eric~W. {Pellegrini}, Ralf~S. {Klessen}, Lena {Maier-Hein}, Carsten {Rother},
  and Ullrich {K{\"o}the}.
\newblock {Analyzing Inverse Problems with Invertible Neural Networks}.
\newblock {\em arXiv e-prints}, page arXiv:1808.04730, Aug 2018.

\bibitem{scale_revnet}
Will {Grathwohl}, Ricky T.~Q. {Chen}, Jesse {Bettencourt}, Ilya {Sutskever},
  and David {Duvenaud}.
\newblock {FFJORD: Free-form Continuous Dynamics for Scalable Reversible
  Generative Models}.
\newblock {\em arXiv e-prints}, page arXiv:1810.01367, Oct 2018.

\bibitem{invertible_ResNet}
Jens {Behrmann}, Will {Grathwohl}, Ricky T.~Q. {Chen}, David {Duvenaud}, and
  J{\"o}rn-Henrik {Jacobsen}.
\newblock {Invertible Residual Networks}.
\newblock {\em arXiv e-prints}, page arXiv:1811.00995, Nov 2018.

\bibitem{behrmann_excess}
Jens {Behrmann}, S{\"o}ren {Dittmer}, Pascal {Fernsel}, and Peter {Maa{\ss}}.
\newblock {Analysis of Invariance and Robustness via Invertibility of
  ReLU-Networks}.
\newblock {\em arXiv e-prints}, page arXiv:1806.09730, Jun 2018.

\bibitem{gilmer_excess}
Justin {Gilmer}, Luke {Metz}, Fartash {Faghri}, Samuel~S. {Schoenholz}, Maithra
  {Raghu}, Martin {Wattenberg}, and Ian {Goodfellow}.
\newblock {Adversarial Spheres}.
\newblock {\em arXiv e-prints}, page arXiv:1801.02774, Jan 2018.

\bibitem{jacobsen_excess}
J{\"o}rn-Henrik {Jacobsen}, Jens {Behrmann}, Richard {Zemel}, and Matthias
  {Bethge}.
\newblock {Excessive Invariance Causes Adversarial Vulnerability}.
\newblock {\em arXiv e-prints}, page arXiv:1811.00401, Nov 2018.

\bibitem{gan}
Ian~J. {Goodfellow}, Jean {Pouget-Abadie}, Mehdi {Mirza}, Bing {Xu}, David
  {Warde-Farley}, Sherjil {Ozair}, Aaron {Courville}, and Yoshua {Bengio}.
\newblock {Generative Adversarial Networks}.
\newblock {\em arXiv e-prints}, page arXiv:1406.2661, Jun 2014.

\bibitem{cgan}
Mehdi {Mirza} and Simon {Osindero}.
\newblock {Conditional Generative Adversarial Nets}.
\newblock {\em arXiv e-prints}, page arXiv:1411.1784, Nov 2014.

\bibitem{cycle_3}
Chris {Donahue}, Zachary~C. {Lipton}, Akshay {Balsubramani}, and Julian
  {McAuley}.
\newblock {Semantically Decomposing the Latent Spaces of Generative Adversarial
  Networks}.
\newblock {\em arXiv e-prints}, page arXiv:1705.07904, May 2017.

\bibitem{cycle_2}
Vincent {Dumoulin}, Ishmael {Belghazi}, Ben {Poole}, Olivier {Mastropietro},
  Alex {Lamb}, Martin {Arjovsky}, and Aaron {Courville}.
\newblock {Adversarially Learned Inference}.
\newblock {\em arXiv e-prints}, page arXiv:1606.00704, Jun 2016.

\bibitem{cycle_4}
Yunfei {Teng}, Anna {Choromanska}, and Mariusz {Bojarski}.
\newblock {Invertible Autoencoder for domain adaptation}.
\newblock {\em arXiv e-prints}, page arXiv:1802.06869, Feb 2018.

\bibitem{cycle_1}
Jun-Yan {Zhu}, Taesung {Park}, Phillip {Isola}, and Alexei~A. {Efros}.
\newblock {Unpaired Image-to-Image Translation using Cycle-Consistent
  Adversarial Networks}.
\newblock {\em arXiv e-prints}, page arXiv:1703.10593, Mar 2017.

\bibitem{kl_divergence}
S.~Kullback and R.~A. Leibler.
\newblock On information and sufficiency.
\newblock {\em Ann. Math. Statist.}, 22(1):79--86, 1951.

\bibitem{mmd}
Arthur Gretton, Karsten~M. Borgwardt, Malte~J. Rasch, Bernhard Sch\"{o}lkopf,
  and Alexander Smola.
\newblock A kernel two-sample test.
\newblock {\em J. Mach. Learn. Res.}, 13:723--773, March 2012.

\bibitem{adam}
Diederik~P. {Kingma} and Jimmy {Ba}.
\newblock {Adam: A Method for Stochastic Optimization}.
\newblock {\em arXiv e-prints}, page arXiv:1412.6980, Dec 2014.

\bibitem{info_loss_2}
Shujian {Yu}, Kristoffer {Wickstr{\o}m}, Robert {Jenssen}, and Jose~C.
  {Principe}.
\newblock {Understanding Convolutional Neural Networks with Information Theory:
  An Initial Exploration}.
\newblock {\em arXiv e-prints}, page arXiv:1804.06537, Apr 2018.

\bibitem{info_loss_1}
Shujian {Yu} and Jose~C. {Principe}.
\newblock {Understanding Autoencoders with Information Theoretic Concepts}.
\newblock {\em arXiv e-prints}, page arXiv:1804.00057, Mar 2018.

\bibitem{info_bottle_3}
Alexander~A. {Alemi}, Ian {Fischer}, Joshua~V. {Dillon}, and Kevin {Murphy}.
\newblock {Deep Variational Information Bottleneck}.
\newblock {\em arXiv e-prints}, page arXiv:1612.00410, Dec 2016.

\bibitem{info_bottle_2}
Ravid {Shwartz-Ziv} and Naftali {Tishby}.
\newblock {Opening the Black Box of Deep Neural Networks via Information}.
\newblock {\em arXiv e-prints}, page arXiv:1703.00810, Mar 2017.

\bibitem{info_bottle_1}
Naftali {Tishby} and Noga {Zaslavsky}.
\newblock {Deep Learning and the Information Bottleneck Principle}.
\newblock {\em arXiv e-prints}, page arXiv:1503.02406, Mar 2015.

\end{thebibliography}

\end{document}